\documentclass[letterpaper]{article} % DO NOT CHANGE THIS
\usepackage{aaai24}  % DO NOT CHANGE THIS
\usepackage{times}  % DO NOT CHANGE THIS
\usepackage{helvet}  % DO NOT CHANGE THIS
\usepackage{courier}  % DO NOT CHANGE THIS
\usepackage[hyphens]{url}  % DO NOT CHANGE THIS
\usepackage{graphicx} % DO NOT CHANGE THIS
\usepackage{natbib}  % DO NOT CHANGE THIS AND DO NOT ADD ANY OPTIONS TO IT
\usepackage{caption} % DO NOT CHANGE THIS AND DO NOT ADD ANY OPTIONS TO IT
\usepackage{algorithm}
\usepackage{algorithmic}

\usepackage{booktabs}
\usepackage{amsmath}
\usepackage{multirow}

\usepackage{newfloat}
\usepackage{listings}
\DeclareCaptionStyle{ruled}{labelfont=normalfont,labelsep=colon,strut=off} % DO NOT CHANGE THIS
\floatstyle{ruled}
\newfloat{listing}{tb}{lst}{}
\floatname{listing}{Listing}
\title{Enhancing Hyperspectral Images via Diffusion Model and Group-Autoencoder Super-resolution Network}
\author {
    Zhaoyang Wang\textsuperscript{\rm 1} \textsuperscript{\rm 2}\thanks{Work done when interning with Dongyang Li at Alibaba.},
    Dongyang Li\textsuperscript{\rm 2} \textsuperscript{\rm 3},
    Mingyang Zhang\textsuperscript{\rm 1}\thanks{Corresponding author.},
    Hao Luo\textsuperscript{\rm 2} \textsuperscript{\rm 3},
    Maoguo Gong\textsuperscript{\rm 1},
}
\affiliations {
    \textsuperscript{\rm 1}Ministry of Education,
Key Laboratory of Collaborative Intelligence Systems, Xidian University\\
    \textsuperscript{\rm 2}DAMO Academy, Alibaba Group, 310023, Hangzhou, China\\
    \textsuperscript{\rm 3}Hupan Lab, 310023, Hangzhou, China\\
    zhaoyangwang@stu.xidian.edu.cn, yingtian.ldy@alibaba-inc.com, myzhang@xidian.edu.cn,michuan.lh@alibaba-inc.com,gong@ieee.org
}

\usepackage{bibentry}
\begin{document}

\maketitle

\begin{abstract}
Existing hyperspectral image (HSI) super-resolution (SR) methods struggle to effectively capture the complex spectral-spatial relationships and low-level details, while diffusion models represent a promising generative model known for their exceptional performance in modeling complex relations and learning high and low-level visual features. 
The direct application of diffusion models to HSI SR is hampered by challenges such as difficulties in model convergence and protracted inference time.
In this work, we introduce a novel Group-Autoencoder (GAE) framework that synergistically combines with the diffusion model to construct a highly effective HSI SR model (DMGASR). Our proposed GAE framework encodes high-dimensional HSI data into low-dimensional latent space where the diffusion model works, thereby alleviating the difficulty of training the diffusion model while maintaining band correlation and considerably reducing inference time.
Experimental results on both natural and remote sensing hyperspectral datasets demonstrate that the proposed method is superior to other state-of-the-art methods both visually and metrically.

\end{abstract}

\section{Introduction}
Hyperspectral images (HSIs) offer plenty of information in the spectral dimension and have been found extensive applications in various fields such as remote sensing \cite{1996Review}, material recognition \cite{2002Invariant}, agriculture \cite{2012Pre}, medical diagnosis \cite{705921b4886f4afaa18d268f6c958727}, and many others. However, the spatial resolution of HSI images is often relatively low due to the
constraints of imaging hardware, which has a negative impact on subsequent HSI applications.
Therefore, the HSI super-resolution (HSI SR) task is critical and meaningful  to enhance the image quality to better serve the subsequent high-level computer vision tasks.

HSI SR can be categorized into two groups depending on whether auxiliary information is required: fusion-based SR \cite{2021Fusformer}, \cite{2020A}
and single-image SR (SISR).
The fusion-based SR methods require additional high-resolution (HR) images as an aid, making them difficult to implement in practical application scenarios and making the SISR methods the research mainstream. 
The current state-of-the-art methods in SISR are predominantly CNN-based \cite{9930808}, \cite{9796466} 
and the majority of existing methods still suffer from modeling complex spectral-spatial relations and characterizing local features and global features comprehensively.

\begin{figure}[t]
\centering
\includegraphics[width=0.95\columnwidth]{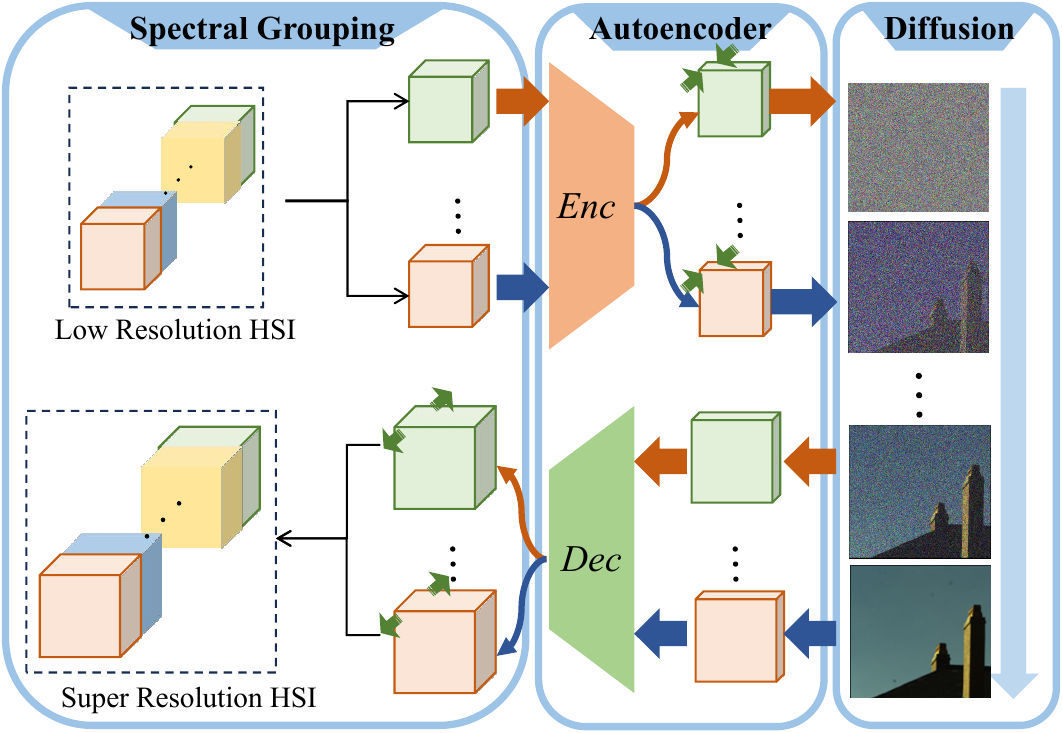} % Reduce the figure size so that it is slightly narrower than the column. Don't use precise values for figure width.This setup will avoid overfull boxes.
\caption{Our proposed framework combines three key techniques: spectral grouping and fusing techniques, autoencoder techniques and diffusion-based SR network.
}
\label{fig1}
%\vspace{-5mm}
\end{figure}

Most recently, as a new emerging record-breaking model, diffusion model \cite{NEURIPS2020_4c5bcfec}, \cite{pmlr-v139-nichol21a} shows superior performance in generation and reconstruction tasks 
and have started to be explored in the field of normal images SR \cite{9887996}, with promising results being reported.  In addition, the diffusion model has been applied to HSI classification \cite{zhou2023hyperspectral}, demonstrating its effectiveness.
The diffusion model shows great ability in acquiring and learning global information and higher-level detailed texture information \cite{zhou2023hyperspectral},
which is ideally suited for handling complex spectral-spatial relationships and capturing global features and local features in dealing with HSI SR problems. The above work has inspired us to investigate the potential of the diffusion model for enhancing HSI SR.

Compared with natural images, HSI data have massive and high dimensional characteristics, and the training samples are not as sufficient as natural images. Directly applying the diffusion model to the HSI SR task results in difficulties in model convergence, while simply performing a band-by-band SR application disrupts spectral continuity and ignores the band similarity, leading to unsatisfactory results as well (see Table \ref{ablation}). Additionally, the requirement for band-by-band SR necessitates multiple inferences, leading to significantly prolonged inference time (see Table \ref{time}).

To address these challenges, we propose a novel SR network for HSI data that integrates diffusion model, autoencoder techniques and spectral grouping techniques, as shown in Figure \ref{fig1}.
We propose a novel autoencoder architecture that can encode a sheet of HSI data into several low-dimensional hidden variables for training the diffusion model. 
By adopting this approach, we address two crucial aspects. 
Firstly, it alleviates the challenges related to training and convergence in diffusion models when facing high spectral dimensionality HSI data and the ``one-to-many" dimensionality reduction effectively reduces information loss during the encoding process, resulting in an enriched abundance of feature information
(see Table \ref{ablation}).
Secondly,  due to the collaborative work of the autoencoders,  our model efficiently narrows down the inference process to a few critical intermediate hidden variables, resulting in a substantial reduction in the inference time (see Table \ref{time}, Figure \ref{balance}), making our approach more efficient and scalable for practical applications in HSI SR tasks.
Our model comprises two primary training stages and following the completion of these training stages, 
the two modules collaborate harmoniously to execute the SR task effectively.

In summary, the main contributions of our work are as follows:
\begin{itemize}
    \item To the best of our knowledge, our work represents the first application of the diffusion model to the field of HSI SR.
We propose a novel diffusion-based SR model that facilitates the implicit capture of both high and low-level features and improves the learning ability of complex spectral-spatial relationships.
    \item We fuse diffusion models with autoencoder techniques to overcome the difficulty of convergence and significantly decrease the inference time in the face of high-dimensional data. 
    \item Extensive experimental results on three publicly available HSI datasets show that our proposed method outperforms state-of-the-art methods in terms of both objective metrics and subjective visual quality.
\end{itemize}

\section{Related Works}
\subsection{Single Hyperspectral Image Super-Resolution}
In recent years, deep convolutional networks have shown impressive capabilities in recovering missing features in HSI data. For instance, in \cite{8499097}, the authors proposed a recursive residual network (GDRRN) that utilizes all data channels as inputs and integrates Spectral Angle Mapper (SAM) into the loss function, which was a groundbreaking advancement. Other researchers, such as \cite{li2020mixed}, presented the Mixed Convolutional Network (MCNet), which employs both 2-D and 3-D convolutions to reduce the computational burden of processing all bands simultaneously. Additionally, \cite{li2021exploring} explored the relationships between 2-D and 3-D convolutions.
In \cite{jiang2020learning}, the Spatial-Spectral Prior Network (SSPSR) was introduced, featuring a group convolution and progressive upsampling framework built upon the spectral grouping strategy. Building on this approach, \cite{9930808} proposed the Group-based Embedding Learning and Integration Network (GELIN), which effectively utilizes information from neighboring spectral bands.
Furthermore, many other research efforts have been based on the spectral grouping strategy, as demonstrated in \cite{liu2022gjtd}, \cite{liu2022cnn}, \cite{liu2022model} and \cite{9380508}. Additionally, some researchers have utilized Transformer-based architectures to learn the complex relationships between spectral and spatial information, as seen in \cite{gao2021stransfuse}, \cite{hu2022fusformer} and \cite{liu2022interactformer}. These works collectively contribute to the advancement of HSI data analysis and SR tasks.

\subsection{Diffusion Based Super-Resolution Model}
In recent years, the diffusion model has demonstrated its remarkable generative capabilities in various domains, including natural image SR tasks \cite{NEURIPS2020_4c5bcfec, pmlr-v139-nichol21a, li2022srdiff, saharia2022image}. Notably, the SR3 model \cite{saharia2022image}, based on the diffusion model, has achieved high-performance results in super-resolving natural images. Stable Diffusion (LDM) \cite{rombach2022high} is another top-performing diffusion method that exhibits exceptional performance in SR tasks.
Moreover, diffusion models have been successfully applied to continuous SR of natural images \cite{Gao_Liu_Zeng_Xu_Li_Luo_Liu_Zhen_Zhang_2023}. Beyond natural images, diffusion models have been found applications in various domains, such as magnetic resonance image SR \cite{mao2023disc, chung2022mr}, and have even been introduced to remote sensing imagery SR \cite{liu2022diffusion}.
The broad applicability and impressive performance of diffusion models showcase their potential for various SR tasks across different domains.

\section{Methodology}
\begin{figure*}[t]
\centering
\includegraphics[width=0.98\textwidth]{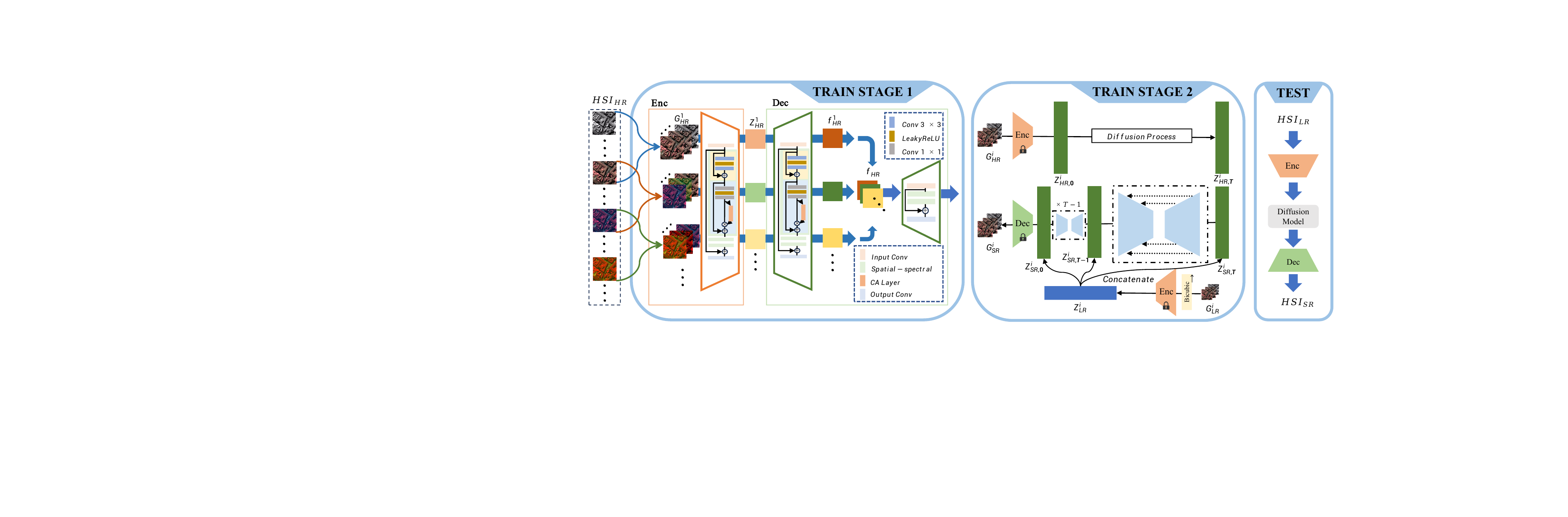} % Reduce the figure size so that it is slightly narrower than the column. Don't use precise values for figure width.This setup will avoid overfull boxes.
\caption{Overview of the proposed model,  In Stage 1, the autoencoder is trained to encode the input data  into a series of hidden variables ($[Z_{HR}^1,Z_{HR}^2 \cdots Z_{HR}^n]$).  
In Stgae 2, the diffusion model is trained. The grouped data  ($G_{HR}^i$ and $G_{LR}^i$) are first encoded, generating hidden variables ($Z_{HR}^i$ and $Z_{LR}^i$) and the $z_{LR}^i$ is added as conditional information by directly concatenating it with the hidden variables ($Z_{SR,t}^i$) at each moment during the denoising process.}
\label{stage}
%\vspace{-5mm}
\end{figure*}

Our model is a two-stage training model that consists of two main parts: the autoencoder and the diffusion SR model, as shown in Figure \ref{stage}. Our proposed model efficiently operates SR on the latent space. Specifically, an autoencoder consisting of an encoder $Enc(\cdot)$ and decoder $Dec(\cdot)$  is trained with a reconstruction objective. Given an HR HSI input $HSI_{HR}$, the encoder $Enc(\cdot)$ maps one single image to several hidden latent variables $[Z^1_{HR},Z^2_{HR} \cdots  Z^n_{HR}]$, and the decoder $Dec(\cdot)$ reconstructs the image from the hidden latent list. In this way, at each timestep $t$, a noisy latent list can be obtained $[Z^1_{HR,t},Z^2_{HR,t} \cdots  Z^n_{HR,t}]$. Beyond the routine training scheme, we also devise conditioning mechanisms to control the SR image by concatenating the LR hidden latent list $[Z^1_{LR},Z^2_{LR} \cdots  Z^n_{LR}]$. In inference, a list-shaped latent is sampled from standard normal distribution $\mathcal{N}(0,1)$ and the denoising model U-Net is used to iteratively remove noise to produce SR latent list. In the end, the SR latent list is passed to the decoder $Dec(\cdot)$ to generate the SR image $HSI_{SR}$.

In the following section, we will introduce the architectural components and principles of the two training stages as well as the detailed testing process.

\subsection{Stage 1: Training the Group Autoencoder}
\textbf{Overall Network Architecture.}
Our autoencoder model, illustrated in Figure \ref{stage}, incorporates a spectral grouping strategy and an asymmetric decoder design. 
Initially, the input data is grouped into bands, and each group is encoded to form a latent variable list. The decoding process is divided into two parts. In the first part, each sub-hidden variable is decoded separately. In the second part, the decoded sub-hidden variables are concatenated and decoded as a whole, resulting in the final decoded image. This two-stage decoding approach allows for effective feature extraction and reconstruction, enhancing the performance of our autoencoder model.

\textbf{Spectral Grouping.}
Compressing all the spectral information directly into low-dimensional space can lead to significant information loss due to the high dimensionality of HSI data \cite{jiang2020learning}. Finding the right compression scale is crucial, as a scale too large may result in excessive information loss, while a scale too small may cause convergence issues during training of the diffusion model.
To strike a balance between information loss and effective coordination with the diffusion model, we introduced the spectral grouping strategy. This strategy involves dividing adjacent bands into groups with certain overlaps between them, allowing the encoder to learn the correlation between bands. 
In this way, we achieve efficient dimensionality reduction of the spectral dimension, reduces information loss, and enhances the richness of the hidden variables' features. Additionally, it considers the similarity between bands, which facilitates a more effective collaboration with the diffusion model. 
This approach enables our model to effectively handle the challenges of HSI SR tasks.

\textbf{Asymmetric Architecture.}
Due to the list-shaped latent variable, we designed an asymmetric decoder with a larger model compared to the encoder to better capture the features. The decoder comprises two primary components: the local decoding part and the global decoding part. In the local decoding part, the variables in each list are initially decoded to decipher the local information features. These decoded variables are then concatenated to match the size of the actual HSI data. Subsequently, the concatenated data is passed through the global decoding part, which decodes the smooth details for the connected parts and enhances the overall effect. This two-stage decoding process allows our model to effectively reconstruct the HSI data and improve feature representation.

\textbf{Loss Function.}
Our loss function consists of four main components: the L1 loss, the spectral angle mapper (SAM) loss, 
the gradient loss and the perceptual loss \cite{johnson2016perceptual}. The $L_1$ loss can be formulated as
\begin{equation}
    L_1 = \frac{1}{N} \sum_{n=1}^{N} ||I_{Re}^n - I_{HR}^n ||
\end{equation}
where $I_{Re}^n$ and $I_{HR}^n$ are the $n$th reconstructed
HSI and original $HSI_{HR}$ and $N$ is the number of images in one training batch.
In addition, to maintain the consistency of the spectral information while performing spatial SR, 
we introduce SAM as part of the loss function, which can be formulated as
\begin{equation}
    L_{SAM} = \frac{1}{N} \sum_{n=1}^{N} \frac{1}{N_p} \sum_{i=1}^{N_p} \frac{1}{\pi} arccos (\frac{I_{Re}^{n,i} \cdot I_{HR}^{n,i}}{||I_{Re}^{n,i}||_2 \cdot ||I_{HR}^{n,i}||_2})
\end{equation}
where $N_p$ is equal to $H \times W$ and $I_{HR}^{n,i}$ refers to the $i$th spectral vector of the $n$th image.
Furthermore, inspired by \cite{9930808}, we add a gradient loss to preserve the sharpness of the reconstructed images in both spatial and spectral domains, which is shown below:
\begin{equation}
    L_{g} = \frac{1}{N} \sum_{n=1}^{N} ||M(I_{HR}^n) - M(I_{Re}^n) ||
\end{equation}
where $M$ computes the gradient value along the horizontal, vertical, and spectral dimensions of the image.
Finally, we add perceptual loss to better take into account the perception of the human visual system, 
to better preserve the high-level features of the images and to better avoid over-smoothing.
The perceptual loss can be formulated as
\begin{equation}
    L_{p} = \frac{1}{N} \sum_{n=1}^{N} ||VGG(I_{HR}^n) - VGG(I_{Re}^n) ||
\end{equation}
where $VGG$ represents the pre-trained VGG19 model \cite{brusilovsky:simonyan2014very}.

In summary, the total loss for our proposed GAE model can be formulated as follows:

\begin{equation}
\text{Loss} = L_1 + \lambda_1 L_{\text{SAM}} + \lambda_2 L_g + \lambda_3 L_p
\end{equation}
In our experiments, we set the weights as $\lambda_1 = 0.3$, $\lambda_2 = 0.1$, and $\lambda_3 = 0.001$ to balance the contributions of the different loss components. This formulation allows our model to effectively optimize and achieve superior results in HSI SR tasks.

\subsection{Stage 2: Training the Diffusion Model}
In our approach, the complete training process of stage 2 is visually represented in Figure \ref{stage}. The GAE plays a key role by simultaneously encoding both $HSI_{HR}$ and $HSI_{LR}$ images, generating separate latent variable lists for each type ($[z_{LR}^1,z_{LR}^2 \cdots z_{LR}^n], [z_{HR}^1,z_{HR}^2 \cdots z_{HR}^n]$). This unique capability allows us to create novel training data pairs by combining corresponding data points from each variable list ($z_{LR}^i,z_{HR}^i$). Subsequently, these paired data points are used to train the diffusion SR model, enabling it to learn and generate high-quality SR results.

For our diffusion SR model, we adopt the architecture from the SR3 framework \cite{saharia2022image}, which has demonstrated excellent performance in natural image SR. By combining the strengths of both the autoencoder and the diffusion model, our proposed approach achieves significant improvements in HSI SR tasks, both qualitatively and quantitatively.
To address the SR task, we incorporate the LR image as conditional information by directly concatenating it with the HR latent image. We choose this straightforward approach over more complex methods like cross-attention mechanisms \cite{rombach2022high} because our conditional information has already been encoded and compressed once. Directly concatenating all the information allows the network to learn more effectively, building upon the demonstrated superiority in the SR3 framework. This effective utilization of conditional information enhances the performance of our model in handling HSI SR tasks.

\subsection{Testing Process}
The complete process of using the trained model for SR tasks is illustrated in Algorithm \ref{test}. First, we encode $HSI_{LR}$ to obtain a list of latent variables ($[z_{LR}^1,z_{LR}^2...z_{LR}^n]$). Next, the diffusion model performs SR on each element inside the list, resulting in a new list of SR latent variables ($[z_{SR}^1,z_{SR}^2...z_{SR}^n]$). Finally, we decode the SR latent variables to obtain the final $HSI_{SR}$ image. This sequential process enables our model to effectively enhance the spatial resolution of HSI data and achieve high-quality SR results.

\begin{algorithm}[t]
    \caption{Testing process}
    \label{test}
    \textbf{Input}: encoder: $Enc(\cdot)$, decoder: $Dec(\cdot)$, noise prediction model: $\epsilon_{\theta}(\cdot)$, LR images: $HSI_{LR}$, time step: $t$ \\
    \textbf{Output}: SR images: $HSI_{SR}$
    \begin{algorithmic}[1] %[1] enables line numbers
    \STATE Encode LR images:
    \[[z_{LR}^1,z_{LR}^2 \cdots z_{LR}^n] = Enc(HSI_{LR})\] 
    \STATE Perform SR on each element in the $[z_{LR}^1,z_{LR}^2...z_{LR}^n]$: \\
    \textbf{for} $i = 0,1 \cdots n$ \textbf{do} \\
    \hspace{0.5em} $z_{T}^i \sim \mathcal{N}(0,I)$ \\
    \hspace{0.5em} \textbf{for} $t = T \cdots 1$ \textbf{do} \\
    \hspace{1.3em} $\epsilon \sim \mathcal{N}(0,I)$ if $t>1$, else $\epsilon = 0$ \\
    \hspace{1.3em}$z_{t-1}^i = \frac{1}{\sqrt{\alpha_t}}(z_{t}^i - \frac{1-\alpha_t}{\sqrt{1-\bar{\alpha_t}}}\epsilon_{\theta}(z_{t}^i,t,z_{LR}^i)) + \sqrt{1-\alpha_t}\epsilon$ \\
     \hspace{0.5em} $z_{SR}^i = z_{0}^i$ \\
    %  \hspace{0.5em} \textbf{end for} \\
    % \textbf{end for}
    \STATE Decode the obtained list $List_{z_{SR}}$: 
    \[HSI_{SR} = Dec([z_{SR}^1,z_{SR}^2 \cdots z_{SR}^n])\]
    \STATE \textbf{Return} $HSI_{SR}$
    \end{algorithmic}
    \end{algorithm}

\section{Experiments}

\begin{table*}[!htbp]
\centering
{\Huge
\resizebox{1\textwidth}{!}{
\begin{tabular}{c  c c c c c c |c c c c c c }
\toprule
\toprule
 \multicolumn{7}{c}{PaviaC Dataset \& Scale = 2}& \multicolumn{6}{c}{PaviaC Dataset \& Scale = 3}\\
\midrule
Models & MPSNR $\uparrow$ & MSSIM $\uparrow$ & CC $\uparrow$ & RMSE $\downarrow$ & SAM $\downarrow$ & ERGAS $\downarrow$ & MPSNR $\uparrow$ & MSSIM $\uparrow$ & CC $\uparrow$ & RMSE $\downarrow$ & SAM $\downarrow$ & ERGAS $\downarrow$\\
\midrule
Bicubic & 30.998 & 0.899 & 0.940 & 0.0292 & 4.675 & 4.567 & 27.865 & 0.785 & 0.877 & 0.0424 & 5.855 & 6.472 \\
EDSR    & 31.342 & 0.904 & 0.940 & 0.0276 & 6.746 & 4.548 & 28.748 & 0.826 & 0.898 & 0.0376 & 7.714 & 5.945 \\
GDRRN   & 31.559 & 0.905 & 0.947 & 0.0271 & 5.186 & 4.324 & 29.172 & 0.835 & 0.908 & 0.0360 & 6.531 & 5.615 \\
SSPSR   & 32.335 & 0.922 & 0.953 & 0.0247 & 5.346 & 4.015 & 29.507 & 0.850 & 0.916 & 0.0346 & 5.987 & 5.416 \\
MCNet   & 32.068 & 0.921 & 0.953 & 0.0262 & 5.219 & 4.174 & 28.193 & 0.802 & 0.889 & 0.0415 & 7.171 & 6.408 \\
CEGATSR & 31.746 & 0.909 & 0.939 & 0.0269 & 5.741 & 4.337 & 28.489 & 0.807 & 0.894 & 0.0392 & 6.816 & 6.051 \\
GELIN   & \underline{33.326} & \underline{0.937} & \underline{0.963} & \underline{0.0222} & \underline{4.099} & \underline{3.553} & \underline{29.611} & \underline{0.850} & \underline{0.918} & \underline{0.0347} & \textbf{5.294} & \underline{5.334} \\
Ours    & \textbf{34.491} & \textbf{0.950} & \textbf{0.971} & \textbf{0.0195} & \textbf{4.080} & \textbf{3.140} & \textbf{30.035} & \textbf{0.867} & \textbf{0.925} & \textbf{0.0328} & \underline{5.715} & \textbf{5.091} \\

\midrule
\midrule
 \multicolumn{7}{c}{Chikusei Dataset \& Scale = 2}& \multicolumn{6}{c}{Chikusei Dataset \& Scale = 3}\\
\midrule

Bicubic & 35.008 & 0.932 & 0.965 & 0.0229 & 1.718 & 3.995 & 31.460 & 0.847 & 0.921 & 0.0345 & 2.547 & 5.935 \\
EDSR    & 35.489 & 0.941 & 0.961 & 0.0198 & 2.444 & 4.525 & 31.962 & 0.868 & 0.925 & 0.0305 & 3.356 & 6.244 \\
GDRRN   & 35.958 & 0.939 & 0.971 & 0.0206 & \underline{1.561} & 3.606 & 32.383 & 0.866 & 0.935 & 0.0305 & \underline{2.398} & 5.402 \\
SSPSR   & 35.723 & 0.944 & 0.965 & 0.0197 & 2.275 & 4.187 & 33.015 & 0.890 & 0.942 & 0.0280 & 2.558 & 5.175 \\
MCNet   & 36.371 & 0.948 & 0.971 & 0.0198 & 1.784 & 3.650 & 32.380 & 0.872 & 0.934 & 0.0309 & 2.496 & 5.581 \\
CEGATSR & 35.866 & 0.938 & 0.957 & 0.0204 & 2.212 & 3.994 & 31.685 & 0.856 & 0.923 & 0.0325 & 3.010 & 5.981 \\
GELIN   & \underline{37.747} & \underline{0.959} & \underline{0.979} & \underline{0.0170} & \textbf{1.384} & \underline{3.011} & \underline{33.796} & \underline{0.900} & \underline{0.952} & \underline{0.0267} & \textbf{2.022} & \underline{4.539} \\
Ours    & \textbf{38.748} & \textbf{0.966} & \textbf{0.982} & \textbf{0.0161} & 1.638 & \textbf{2.738} & \textbf{34.192} & \textbf{0.909} & \textbf{0.954} & \textbf{0.0264} & 2.637 & \textbf{4.364} \\

\midrule
\midrule
 \multicolumn{7}{c}{Harvard Dataset \& Scale = 2}& \multicolumn{6}{c}{Harvard Dataset \& Scale = 3}\\
\midrule

Bicubic & 44.813 & 0.972 & 0.974 & 0.00840 & 2.623 & 3.183 & 41.717 & 0.946 & 0.951 & 0.01230 & 3.081 & 4.446 \\
EDSR    & 44.945 & 0.976 & 0.972 & 0.00714 & 3.588 & 3.376 & 43.102 & 0.957 & 0.957 & 0.00994 & 3.687 & 4.024 \\
GDRRN   & 45.868 & 0.975 & 0.976 & 0.00748 & \underline{2.556} & 2.927 & 43.129 & 0.952 & 0.959 & 0.01060 & \textbf{2.970} & 3.949 \\
SSPSR   & 44.939 & 0.977 & 0.975 & 0.00737 & 3.489 & 3.433 & 43.409 & \underline{0.959} & 0.960 & \underline{0.00974} & 3.478 & 3.868 \\
MCNet   & 46.367 & 0.973 & 0.939 & 0.00713 & 3.300 & 3.738 & 42.745 & 0.952 & 0.953 & 0.01060 & 3.475 & 4.463 \\
GELIN   & \textbf{47.024} & \textbf{0.981} & 0.981 & \textbf{0.00623} & \textbf{2.530} & \underline{2.576} & \underline{43.653} & 0.957 & \underline{0.961} & 0.00992 & \underline{3.000} &\underline{ 3.765} \\
Ours    & \underline{46.953} & \underline{0.979} & \textbf{0.982} & \underline{0.00678} & 3.079 & \textbf{2.527} & \textbf{44.028} & \textbf{0.959} & \textbf{0.966} & \textbf{0.00965} & 3.596 & \textbf{3.525} \\
\bottomrule
\bottomrule

\end{tabular}}
}
\caption{Quantitative results on the PaviaC dataset, Chikusei dataset and Harvard dataset at different scales. Bold represents the best result and underline represents the second best.}
\label{Total1}
\vspace{-4mm}
\end{table*}

\begin{table*}[!htbp]
\centering
{\small
% \resizebox{1\textwidth}{!}{
\begin{tabular}{lc c c c c c}
\toprule
\toprule
 \multicolumn{7}{c}{PaviaC Dataset \& Scale = 4}\\
\midrule
              Models 
&MPSNR $\uparrow$ & MSSIM $\uparrow$ & CC $\uparrow$ & RMSE $\downarrow$ & SAM $\downarrow$ & ERGAS $\downarrow$\\
\midrule
              Bicubic 
&26.270 & 0.678 & 0.820 & 0.0514 & 6.623 & 7.762 \\
              EDSR    
&27.098 & 0.741 & 0.851 & 0.0458 & 8.236 & 7.130 \\
              GDRRN   
&27.445 & 0.749 & 0.864 & 0.0443 & 6.557 & 6.822 \\
              SSPSR   
&27.768 & 0.771 & 0.876 & 0.0428 & \underline{6.320} & 6.562 \\
              MCNet   
&\underline{27.854} & 0.770 & \underline{0.877} & \underline{0.0424} & 6.302 & \underline{6.504} \\
              CEGATSR 
&27.278 & 0.730 & 0.860 & 0.0454 & 6.425 & 6.934 \\
              GELIN   
&27.592 & \underline{0.770} & 0.870 & 0.0439 & \textbf{6.265} & 6.679 \\
              Ours    
&\textbf{27.928} & \textbf{0.785} & \textbf{0.880} & \textbf{0.0423} & 7.406 & \textbf{6.428} \\
\midrule
\midrule
 \multicolumn{7}{c}{Chikusei Dataset \& Scale = 4}\\
\midrule

              Bicubic 
&29.676 & 0.770 & 0.882 & 0.0425 & 3.161 & 7.275 \\
              EDSR    
&29.976 & 0.799 & 0.893 & 0.0386 & 4.127 & 7.547 \\
              GDRRN   
&30.658 & 0.801 & 0.905 & 0.0374 & \underline{2.913} & 6.551 \\
              SSPSR   
&30.858 & 0.823 & 0.914 & 0.0355 & 3.196 & 6.651 \\
              MCNet   
&31.189 & 0.821 & 0.916 & \underline{0.0354} & 2.955 & 8.284 \\
              CEGATSR 
&30.569 & 0.806 & 0.908 & 0.0374 & 3.082 & 6.757 \\
              GELIN   
&\underline{31.095} & \underline{0.838} & \underline{0.914} & 0.0366 & \textbf{2.834} & \underline{6.102} \\
              Ours    
&\textbf{32.248} & \textbf{0.860} & \textbf{0.929} & \textbf{0.0332} & 3.507 & \textbf{5.378} \\

\midrule
\midrule
 \multicolumn{7}{c}{Harvard Dataset \& Scale = 4}\\
\midrule

              Bicubic 
&39.940 & 0.926 & 0.934 & 0.0152 & 3.345 & 5.354 \\
              EDSR    
&41.330 & 0.940 & 0.945 & 0.0119 & 4.039 & 4.748 \\
              GDRRN   
&40.369 & 0.933 & 0.922 & 0.0127 & 4.120 & 5.866 \\
              SSPSR   
&41.929 & 0.941 & 0.952 & 0.0115 & 3.513 & 4.410 \\
              MCNet   
&41.986 & 0.939 & 0.947 & 0.0120 & \underline{3.333} & 4.482 \\
             GELIN   
&\underline{42.673} & \underline{0.945} & \underline{0.959} & \underline{0.0110} & \textbf{3.156} & \underline{4.032} \\
              Ours    &\textbf{43.132} & \textbf{0.948} & \textbf{0.961} & \textbf{0.0109} & 3.534 & \textbf{3.883} \\
\bottomrule
\bottomrule

\end{tabular}}
% }
\caption{Quantitative results on the PaviaC dataset, Chikusei dataset and Harvard dataset at different scales. Bold represents the best result and underline represents the second best.}
\label{Total2}
\vspace{-2mm}
\end{table*}

\begin{figure*}[h]
\centering
\includegraphics[width=0.95\textwidth]{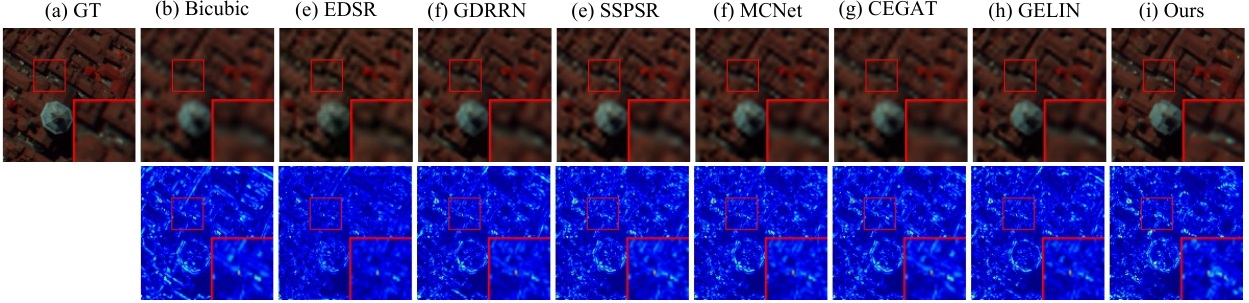} % Reduce the figure size so that it is slightly narrower than the column. Don't use precise values for figure width.This setup will avoid overfull boxes.
\caption{Qualitative results of different models at scale 4 with the corresponding error maps of the PaviaC dataset. The false-color image is used for clear visualization (red: 100, green: 30, and blue: 10).}
\label{Pav}
% \vspace{-3mm}
\end{figure*}

\begin{figure*}[!htbp]
\centering
\includegraphics[width=0.95\textwidth]{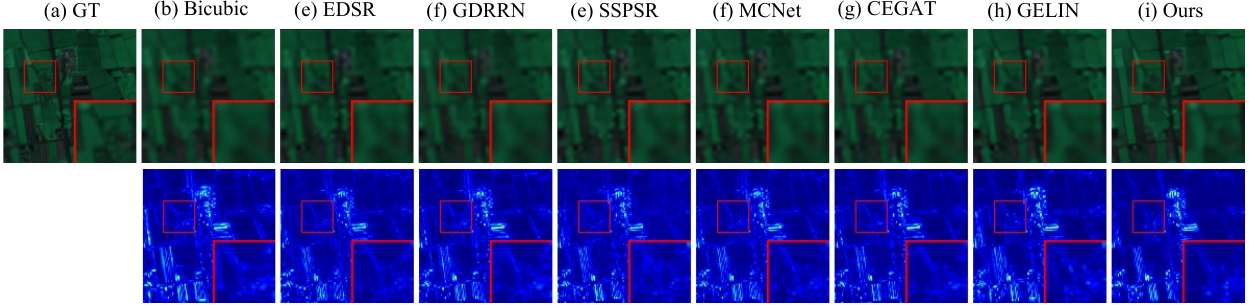} % Reduce the figure size so that it is slightly narrower than the column. Don't use precise values for figure width.This setup will avoid overfull boxes.
\caption{Qualitative results of different models at scale 4 with the corresponding error maps of the Chikusei dataset. The false-color image is used for clear visualization (red: 70, green: 100, and blue: 36).}
\label{Chi}
% \vspace{-3mm}
\end{figure*}

\begin{figure*}[!htbp]
\centering
\includegraphics[width=0.95\textwidth]{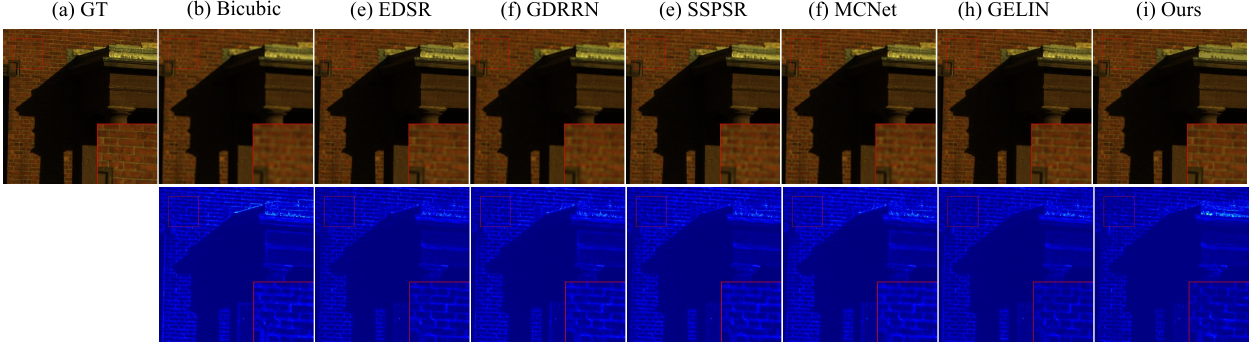} % Reduce the figure size so that it is slightly narrower than the column. Don't use precise values for figure width.This setup will avoid overfull boxes.
\caption{Qualitative results of different models at scale 4 with the corresponding error maps of the Harvard dataset. The false-color image is used for clear visualization (red: 25, green: 15, and blue: 2).}
\label{Har}
% \vspace{-3mm}
\end{figure*}

\subsection{Dataset}
In our experiments, we used three publicly available datasets to validate the performance of our model. These datasets include two remote-sensing HSI datasets: Pavia Center (PaviaC) dataset and Chikusei dataset\cite{unknown}, and one natural image HSI dataset: Harvard dataset \cite{5995660}. These datasets were selected to cover different scenarios and challenges for HSI SR tasks.

\subsection{Evaluation Measures}
In our comprehensive experiments, we employed six widely-used evaluation indices: Peak Signal-to-Noise Ratio (PSNR), Spectral Angle Mapper (SAM), Structural Similarity (SSIM), Cross Correlation (CC), Relative Dimensionless Global Error in Synthesis (ERGAS), and Root-Mean-Squared Error (RMSE). For PSNR and SSIM, we calculated their mean values over all spectral bands. The best values for these indices are $+\infty$, 0, 1, 1, 0, and 0, respectively.

\subsection{Implementation Details}
We used the Adam optimizer with $\beta_1 =0.9$ and $\beta_2 =0.999$ for training, with a batch size of 8 for the Harvard dataset and 4 for the PaviaC and Chikusei datasets. The learning rate was set to $1e^{-4}$ during GAE training and reduced to $1e^{-5}$ for the diffusion model. During the training process, we utilized a pre-trained SR3 diffusion model. In the GAE module, bands were divided into subgroups of size 16 for PaviaC and Chikusei datasets, and 8 for the Harvard dataset, with one-quarter overlap between subgroups. 
% The diffusion model employed a cosine schedule with 20 training steps and a dropout rate of 0.2 for all experiments.

\subsection{Results and Comparison with SOTA}
To validate the effectiveness of our model, we conducted a comprehensive comparison with various state-of-the-art SISR methods. We used standard bicubic interpolation as a baseline and evaluated our results against six SOTA SISR methods: EDSR \cite{Lim_Son_Kim_Nah_Lee_2017}, GDRRN \cite{8499097}, SSPSR \cite{jiang2020learning}, MCNet \cite{li2020mixed}, CEGATSR \cite{liu2022cnn}, and GELIN \cite{9930808}.
This comparison was conducted on three different datasets at three different scales. However, due to network design and CUDA memory limitations, the CEGATSR network could not be compared on the Harvard dataset. Nonetheless, we made every effort to reproduce the performance of each compared network to ensure a fair evaluation.

\textbf{Quantitative Experimental Results.}
It can be noticed that our model achieves the best performance on the PaviaC dataset, outperforming other methods in various evaluation metrics. We further validated its effectiveness on the Chikusei and Harvard datasets, where it consistently demonstrated excellent results in most metrics, as shown in Table \ref{Total1} and Table \ref{Total2}. Although the SAM index improvement was less significant, which may be attributed to the segmented data processing caused by the two-stage training approach, our model still showed superiority in other evaluation indicators. Overall, it achieved an approximate improvement of 0.5-0.8 dB in PSNR compared to other methods, confirming its effectiveness in HSI SR tasks.

\textbf{Qualitative Experimental Results.}
Our model exhibits remarkable superiority in SR results, as demonstrated in Figure \ref{Pav} for the PaviaC dataset, Figure \ref{Chi} for the Chikusei dataset, and Figure \ref{Har} for the Harvard dataset. The visual comparisons and error maps vividly showcase our model's ability to preserve and enhance texture details, resulting in clearer and more accurate reconstructions compared to other methods. The outstanding visual performance aligns with the superior metric results, solidifying the effectiveness of our approach for HSI SR tasks.

\begin{figure*}[!htbp]
    \centering
    \includegraphics[width=1\linewidth]{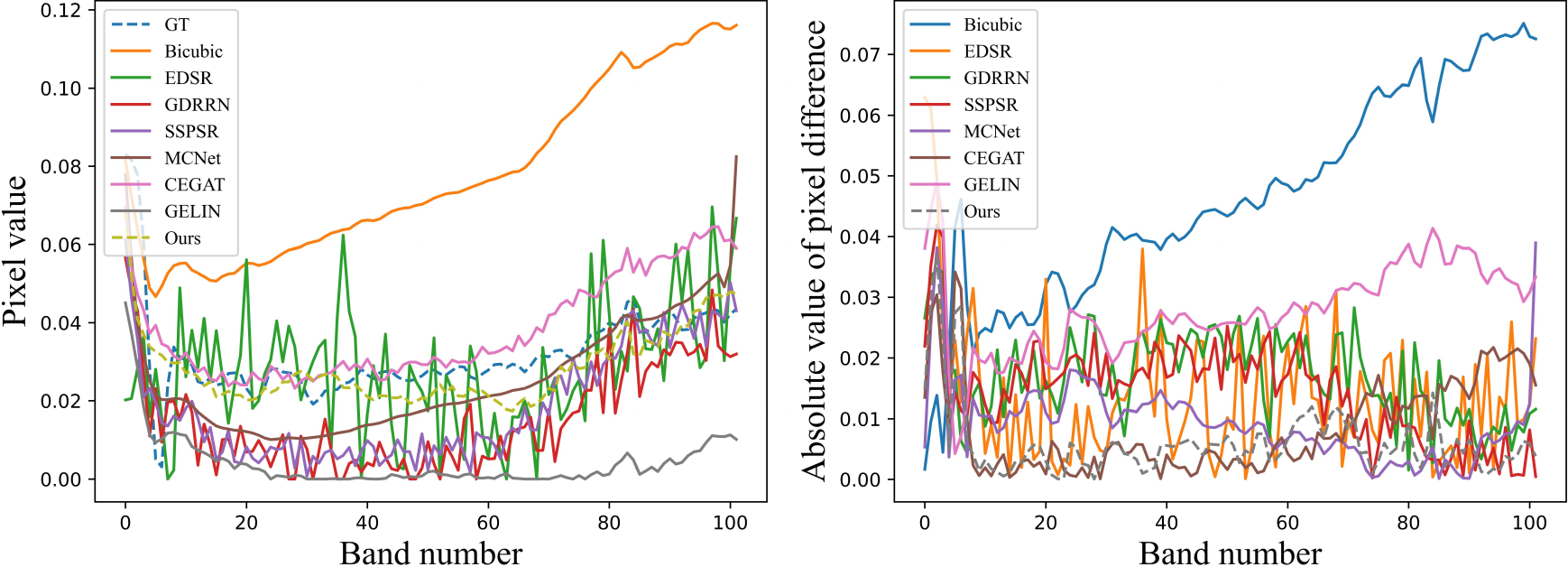}
    \caption{Example spectral curves and difference curves on a selected
pixel value of PaviaC datasets with a scale factor of 4.}
    \label{char1}
    \vspace{-5mm}
\end{figure*}

\begin{figure*}[!htbp]
    \centering
    \includegraphics[width=1\linewidth]{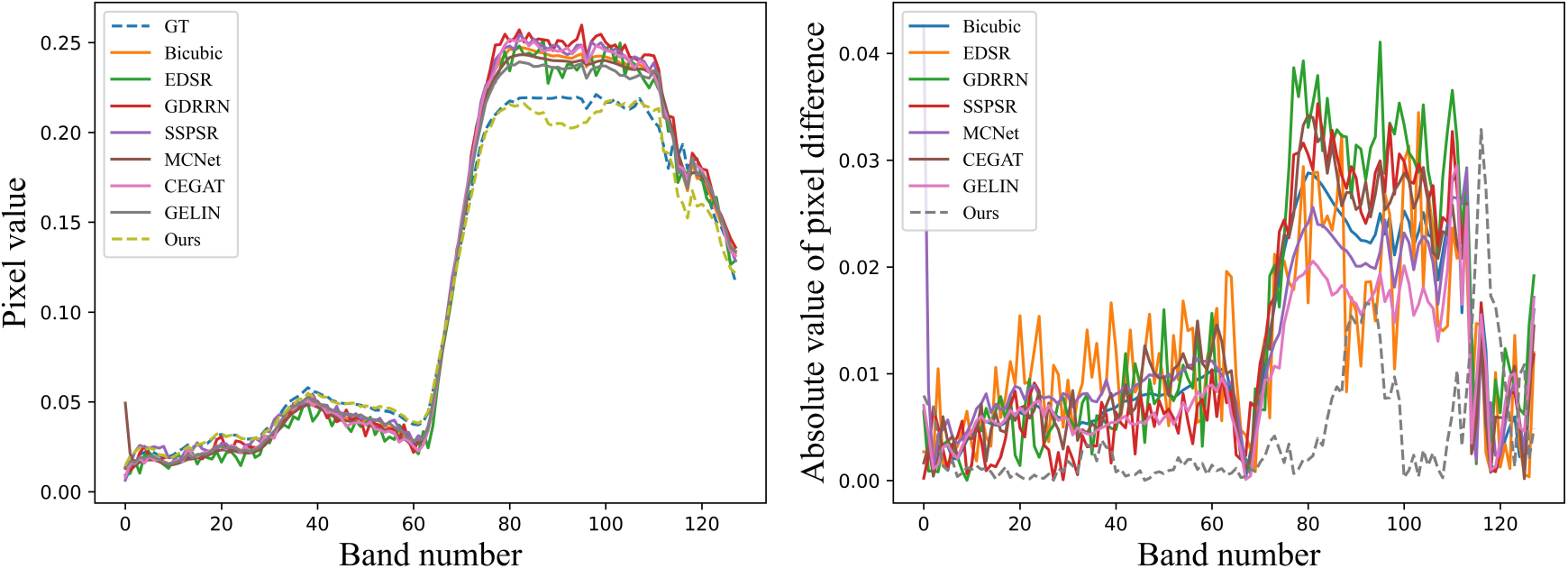}
    \caption{Example spectral curves and difference curves on a selected
pixel value of Chikusei datasets with a scale factor of 4.}
    \label{char2}
    \vspace{-5mm}
\end{figure*}

\begin{figure*}[!htbp]
    \centering
    \includegraphics[width=1\linewidth]{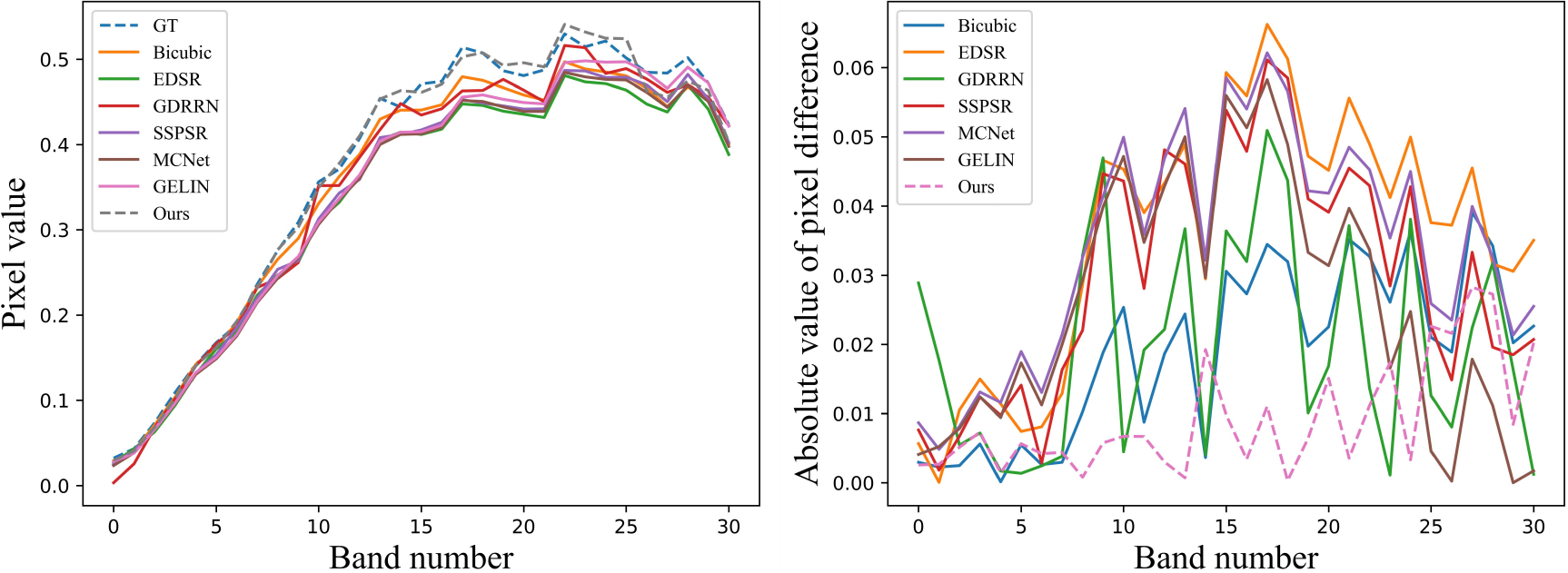}
    \caption{Example spectral curves and difference curves on a selected
pixel value of Harvard datasets with a scale factor of 4.}
    \label{char3}
    \vspace{-5mm}
\end{figure*}

\textbf{Spectral Distortion Comparison.}
Finally, we also compared the spectral distortion among the different methods. As shown in Figure \ref{char1}, Figure \ref{char2} and Figure \ref{char3}, the HSIs reconstructed by our proposed method exhibit the highest spectral fidelity, with minimal spectral distortion. The ability to maintain high spectral fidelity is crucial for HSI tasks, and our model excels in this aspect, making it a robust and reliable solution for HSI SR problems.

\begin{table*}[!htbp]
\centering
{\small
\begin{tabular}{c |c c c c c c}
\toprule
\toprule
Models  & MPSNR $\uparrow$ & MSSIM $\uparrow$ & CC $\uparrow$ & RMSE $\downarrow$ & SAM $\downarrow$ & ERGAS $\downarrow$ \\
\midrule
\multicolumn{7}{c}{$Scale=2$} \\
\midrule
Ours & 34.491 & 0.950 & 0.971 & 0.0195 & 4.080 & 3.140 \\
Ours - w/o GD & 33.761 & 0.946 & 0.966 & 0.0214 & 4.714 & 3.410 \\
Ours - w/o GS & 32.442 & 0.929 & 0.954 & 0.0246 & 5.626 & 4.021 \\
Diff -w PB & 31.319 & 0.827 & 0.878 & 0.0277 & 6.715 & 5.122 \\
Diff -w FB & 10.948 & 0.043 & 0.084 & 0.3501 & 51.461 & 56.450 \\

\midrule
\multicolumn{7}{c}{$Scale=3$} \\
\midrule
Ours & 30.035 & 0.867 & 0.925 & 0.0328 & 5.715 & 5.091 \\
Ours - w/o GD & 29.685 & 0.862 & 0.919 & 0.0341 & 6.442 & 5.296 \\
Ours - w/o GS & 29.469 & 0.854 & 0.915 & 0.0347 & 6.073 & 5.467 \\
Diff -w PB & 29.172 & 0.756 & 0.833 & 0.0353 & 7.383 & 6.202 \\
Diff -w FB & 10.920 & 0.038 & 0.065 & 0.3483 & 53.411 & 54.513 \\

\midrule
\multicolumn{7}{c}{$Scale=4$} \\
\midrule
Ours & 27.928 & 0.785 & 0.880 & 0.0423 & 7.406 & 6.428 \\
Ours - w/o GD & 27.624 & 0.778 & 0.874 & 0.0436 & 8.227 & 6.678 \\
Ours - w/o GS & 27.432 & 0.759 & 0.867 & 0.0433 & 7.193 & 6.822 \\
Diff -w PB & 27.262 & 0.757 & 0.867 & 0.0451 & 10.492 & 6.969 \\
Diff -w FB & 10.906 & 0.043 & 0.088 & 0.3521 & 53.061 & 57.968 \\
\bottomrule
\bottomrule
\end{tabular}}
\caption{Ablation results of quantitative performance on the PaviaC dataset at scale 2,3,4.
GD: global decoding, GS: spectral grouping strategy, PB: without GAE and trained with partial bands step by step,
FB: without GAE and trained with full-bands.
}
\label{ablation}
\end{table*}
\subsection{Ablation Study}
We conducted ablation experiments to assess the effectiveness of each component in our model. Table \ref{ablation} presents the results, where we evaluated the impact of the asymmetric autoencoder, the spectral grouping strategy, and the overall performance of our autoencoder. These experiments allowed us to verify the significance of each component in contributing to the overall performance of our model.

\textbf{Asymmetric Structure.}
The design of the asymmetric autoencoder was essential to accommodate the list-shaped latent variable in our model. As shown in Table \ref{ablation}, removing the global decoding part led to a significant decrease in results, highlighting the crucial role of this structure in our model's effectiveness.

\textbf{Spectral Grouping.}
The spectral dimension grouping strategy effectively reduces unnecessary computational costs by grouping similar spectral bands together, allowing the model to more efficiently utilize the spectral information of the image leading to effective coordination between the autoencoder and the diffusion model. As evident in Table \ref{ablation}, the results without the spectral grouping strategy were even worse, further underscoring its significance in achieving successful and improved results in our model.

\textbf{Autoencoder.}
Removing the autoencoder module and directly using the diffusion SR model resulted in poor performance due to the large amount of redundant data in HSI and limited training samples. Training with partial bands step by step led to fragmented learning, resulting in inferior performance as well. The autoencoder plays a crucial role in handling the rich HSI information and facilitating effective encoding for superior results with the diffusion SR model.

\begin{table}[t]
    \centering
    \small
    \begin{tabular}{c |c c c }
    \toprule
    Settings  & MPSNR $\uparrow$ & MSSIM $\uparrow$  & ERGAS $\downarrow$  \\
    \midrule
    $n_{subs} = 12, n_{ovls} = 4$ &27.784  &0.784&6.537  \\
    $n_{subs} = 16, n_{ovls} = 4$ &\textbf{27.928}  &\textbf{0.785} &\textbf{6.428 } \\
    $n_{subs} = 24, n_{ovls} = 6$ &27.759  &0.781 &6.563  \\
    $n_{subs} = 32, n_{ovls} = 8$ &27.736  &0.778 &6.596  \\
    \bottomrule
    \end{tabular}
    % }
    \caption{Ablation results for different settings of the number of subgroups bands ($n_{subs}$) and the number of overlaps bands ($n_{ovls}$) on the PaviaC dataset at scale 4.
    }
    \label{group}

    \end{table}
\begin{figure}[t]
    \centering
    \includegraphics[width=\columnwidth]{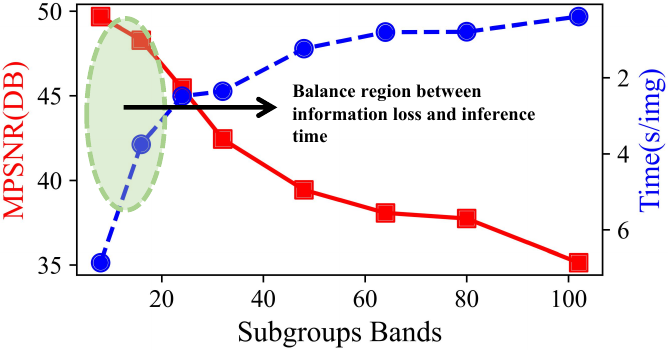}
    \caption{Comparison results of information loss
    and inference time on different subgroups numbers.}
    \label{balance}
\end{figure}

\subsection{Comparison of Subgroups Numbers}
We provide two types of evidence to support the rationality of our subgroups selection. Firstly, we present the overall SR performance of the model under different subgroups bands settings, where we compare the results in terms of several commonly used evaluation metrics. The results are shown in Table \ref{group}. Secondly, we evaluate the information reconstruction loss (measured by PSNR) and the inference time under different subgroups bands settings, as shown in Figure \ref{balance}. It can be observed that our selection not only achieves high SR performance, but also strikes the right balance between information loss and inference time.

\begin{table}[t]
    \centering
    \small

    \begin{tabular}{c | c c c c }
    \toprule
    Models  &  MPSNR   & $T_{64}$ (s/img) & $T_{128}$ (s/img)  & $T_{256}$(s/img) \\
    \midrule
    Ours & 27.928 &3.752 &3.876 & 6.174\\
    Diff-PB & 27.262 & 13.629 &14.124 &  23.018 \\
    Diff-FB & 10.906 &0.396 &0.435 & 0.914\\
    \bottomrule
    \end{tabular}

    \caption{Comparison results of inference times on the PaviaC dataset. $T_{64}$ represents the average inference time of  $64 \times 64$ images. 
    % Diff-PB: without GAE and tested with partial bands step by step, Diff-FB: without GAE and tested with full-bands.
    }
    \label{time}

    \end{table}

\subsection{Comparison of Inference Time}
We also conducted experiments to compare the inference time with the pure diffusion model (without our GAE), as presented in Table \ref{time}. Our method demonstrates a remarkable reduction in the inference time while achieving superior performance. It can be observed that directly applying the diffusion model for full-band SR results in poor performance, despite the advantage of shorter inference time. On the other hand, performing SR on each band separately using the diffusion model requires traversing the entire band spectrum, resulting in lengthy inference steps and ignoring the spectral correlations in HSI data, which in turn leads to suboptimal results. In contrast, our approach significantly reduces the number of inference steps by performing SR on only a few intermediate hidden variables, resulting in much faster inference. Furthermore, the integration of the GAE takes into account the spectral similarity, which further enhances the SR results.

Overall, our method not only achieves better performance in terms of SR quality but also dramatically reduces the inference time, making it a highly efficient and effective solution for HSI SR tasks.

\section{Conclusion}
In this paper, we proposed a novel two-stage diffusion-based framework for HSI SR tasks. Our approach effectively addresses the challenge of the diffusion model's convergence with high-dimensional data and significantly reduces the inference time by integrating a autoencoder with the diffusion model. This combination enables efficient computation and facilitates superior SR results.
and achieves significant improvements both visually and metrically. Going forward, we plan to explore more applications of diffusion models in HSI tasks, aiming to further enhance the field of HSI research.

\section{Acknowledgments}
The authors gratefully acknowledge the support from the National Natural Science Foundation of China under Grant No. 62376205 and Grant No. 62036006, as well as the support from Alibaba Group through Alibaba Innovative Research Program. Their contributions have been instrumental to the success of this research; the views and findings herein are those of the authors and do not necessarily reflect the views of the funders.
% \bigskip
% \noindent Thank you for reading these instructions carefully. We look forward to receiving your electronic files!

% \bibliography{aaai24}

% \appendix
% \clearpage 

\bibliography{aaai24}

\begin{thebibliography}{35}
\providecommand{\natexlab}[1]{#1}

\bibitem[{Chakrabarti and Zickler(2011)}]{5995660}
Chakrabarti, A.; and Zickler, T. 2011.
\newblock Statistics of real-world hyperspectral images.
\newblock In \emph{CVPR 2011}, 193--200.

\bibitem[{Chung, Lee, and Ye(2022)}]{chung2022mr}
Chung, H.; Lee, E.~S.; and Ye, J.~C. 2022.
\newblock MR image denoising and super-resolution using regularized reverse diffusion.
\newblock \emph{IEEE Transactions on Medical Imaging}, 42(4): 922--934.

\bibitem[{Cloutis(1996)}]{1996Review}
Cloutis, E.~A. 1996.
\newblock Review Article Hyperspectral geological remote sensing: evaluation of analytical techniques.
\newblock \emph{International Journal of Remote Sensing}, 17(12): 2215--2242.

\bibitem[{Fei(2020)}]{705921b4886f4afaa18d268f6c958727}
Fei, B. 2020.
\newblock \emph{Hyperspectral imaging in medical applications}, 523--565.
\newblock Data Handling in Science and Technology. Elsevier Ltd.

\bibitem[{Gao et~al.(2021)Gao, Liu, Yang, Chen, Wan, Xiao, and Qian}]{gao2021stransfuse}
Gao, L.; Liu, H.; Yang, M.; Chen, L.; Wan, Y.; Xiao, Z.; and Qian, Y. 2021.
\newblock STransFuse: Fusing swin transformer and convolutional neural network for remote sensing image semantic segmentation.
\newblock \emph{IEEE Journal of Selected Topics in Applied Earth Observations and Remote Sensing}, 14: 10990--11003.

\bibitem[{Gao et~al.(2023)Gao, Liu, Zeng, Xu, Li, Luo, Liu, Zhen, and Zhang}]{Gao_Liu_Zeng_Xu_Li_Luo_Liu_Zhen_Zhang_2023}
Gao, S.; Liu, X.; Zeng, B.; Xu, S.; Li, Y.; Luo, X.; Liu, J.; Zhen, X.; and Zhang, B. 2023.
\newblock Implicit Diffusion Models for Continuous Super-Resolution.

\bibitem[{Ho, Jain, and Abbeel(2020)}]{NEURIPS2020_4c5bcfec}
Ho, J.; Jain, A.; and Abbeel, P. 2020.
\newblock Denoising Diffusion Probabilistic Models.
\newblock In Larochelle, H.; Ranzato, M.; Hadsell, R.; Balcan, M.; and Lin, H., eds., \emph{Advances in Neural Information Processing Systems}, volume~33, 6840--6851. Curran Associates, Inc.

\bibitem[{Hu, Huang, and Deng(2021)}]{2021Fusformer}
Hu, J.~F.; Huang, T.~Z.; and Deng, L.~J. 2021.
\newblock Fusformer: A Transformer-based Fusion Approach for Hyperspectral Image Super-resolution.
\newblock \emph{IEEE Transactions on Geoscience and Remote Sensing}.

\bibitem[{Hu et~al.(2022)Hu, Huang, Deng, Dou, Hong, and Vivone}]{hu2022fusformer}
Hu, J.-F.; Huang, T.-Z.; Deng, L.-J.; Dou, H.-X.; Hong, D.; and Vivone, G. 2022.
\newblock Fusformer: A transformer-based fusion network for hyperspectral image super-resolution.
\newblock \emph{IEEE Geoscience and Remote Sensing Letters}, 19: 1--5.

\bibitem[{Jiang et~al.(2020)Jiang, Sun, Liu, and Ma}]{jiang2020learning}
Jiang, J.; Sun, H.; Liu, X.; and Ma, J. 2020.
\newblock Learning spatial-spectral prior for super-resolution of hyperspectral imagery.
\newblock \emph{IEEE Transactions on Computational Imaging}, 6: 1082--1096.

\bibitem[{Johnson, Alahi, and Fei-Fei(2016)}]{johnson2016perceptual}
Johnson, J.; Alahi, A.; and Fei-Fei, L. 2016.
\newblock Perceptual Losses for Real-Time Style Transfer and Super-Resolution.

\bibitem[{Kersting et~al.(2012)Kersting, Xu, Wahabzada, Bauckhage, Thurau, Roemer, Ballvora, Rascher, Léon, and Pluemer}]{2012Pre}
Kersting, K.; Xu, Z.; Wahabzada, M.; Bauckhage, C.; Thurau, C.; Roemer, C.; Ballvora, A.; Rascher, U.; Léon, J.; and Pluemer, L. 2012.
\newblock Pre-symptomatic prediction of plant drought stress using dirichlet-aggregation regression on hyperspectral images.
\newblock In \emph{National Conference on Artificial Intelligence}.

\bibitem[{Li et~al.(2022)Li, Yang, Chang, Chen, Feng, Xu, Li, and Chen}]{li2022srdiff}
Li, H.; Yang, Y.; Chang, M.; Chen, S.; Feng, H.; Xu, Z.; Li, Q.; and Chen, Y. 2022.
\newblock Srdiff: Single image super-resolution with diffusion probabilistic models.
\newblock \emph{Neurocomputing}, 479: 47--59.

\bibitem[{Li, Wang, and Li(2020)}]{li2020mixed}
Li, Q.; Wang, Q.; and Li, X. 2020.
\newblock Mixed 2D/3D convolutional network for hyperspectral image super-resolution.
\newblock \emph{Remote sensing}, 12(10): 1660.

\bibitem[{Li, Wang, and Li(2021)}]{li2021exploring}
Li, Q.; Wang, Q.; and Li, X. 2021.
\newblock Exploring the relationship between 2D/3D convolution for hyperspectral image super-resolution.
\newblock \emph{IEEE Transactions on Geoscience and Remote Sensing}, 59(10): 8693--8703.

\bibitem[{Li et~al.(2018)Li, Zhang, Dingl, Wei, and Zhang}]{8499097}
Li, Y.; Zhang, L.; Dingl, C.; Wei, W.; and Zhang, Y. 2018.
\newblock Single Hyperspectral Image Super-Resolution with Grouped Deep Recursive Residual Network.
\newblock In \emph{2018 IEEE Fourth International Conference on Multimedia Big Data (BigMM)}, 1--4.

\bibitem[{Lim et~al.(2017)Lim, Son, Kim, Nah, and Lee}]{Lim_Son_Kim_Nah_Lee_2017}
Lim, B.; Son, S.; Kim, H.; Nah, S.; and Lee, K.~M. 2017.
\newblock Enhanced Deep Residual Networks for Single Image Super-Resolution.
\newblock In \emph{2017 IEEE Conference on Computer Vision and Pattern Recognition Workshops (CVPRW)}.

\bibitem[{Liu and Dong(2022)}]{liu2022cnn}
Liu, C.; and Dong, Y. 2022.
\newblock CNN-Enhanced graph attention network for hyperspectral image super-resolution using non-local self-similarity.
\newblock \emph{International Journal of Remote Sensing}, 43(13): 4810--4835.

\bibitem[{Liu, Fan, and Zhang(2022)}]{liu2022gjtd}
Liu, C.; Fan, Z.; and Zhang, G. 2022.
\newblock Gjtd-lr: A trainable grouped joint tensor dictionary with low-rank prior for single hyperspectral image super-resolution.
\newblock \emph{IEEE Transactions on Geoscience and Remote Sensing}, 60: 1--17.

\bibitem[{Liu et~al.(2020)Liu, Wu, Xiao, Sun, and Yan}]{2020A}
Liu, J.; Wu, Z.; Xiao, L.; Sun, J.; and Yan, H. 2020.
\newblock A Truncated Matrix Decomposition for Hyperspectral Image Super-Resolution.
\newblock \emph{IEEE Transactions on Image Processing}, PP(99): 1--1.

\bibitem[{Liu et~al.(2022{\natexlab{a}})Liu, Wu, Xiao, and Wu}]{liu2022model}
Liu, J.; Wu, Z.; Xiao, L.; and Wu, X.-J. 2022{\natexlab{a}}.
\newblock Model inspired autoencoder for unsupervised hyperspectral image super-resolution.
\newblock \emph{IEEE Transactions on Geoscience and Remote Sensing}, 60: 1--12.

\bibitem[{Liu et~al.(2022{\natexlab{b}})Liu, Yuan, Pan, Fu, Liu, and Lu}]{liu2022diffusion}
Liu, J.; Yuan, Z.; Pan, Z.; Fu, Y.; Liu, L.; and Lu, B. 2022{\natexlab{b}}.
\newblock Diffusion model with detail complement for super-resolution of remote sensing.
\newblock \emph{Remote Sensing}, 14(19): 4834.

\bibitem[{Liu et~al.(2022{\natexlab{c}})Liu, Hu, Kang, Luo, and Fan}]{9796466}
Liu, Y.; Hu, J.; Kang, X.; Luo, J.; and Fan, S. 2022{\natexlab{c}}.
\newblock Interactformer: Interactive Transformer and CNN for Hyperspectral Image Super-Resolution.
\newblock \emph{IEEE Transactions on Geoscience and Remote Sensing}, 60: 1--15.

\bibitem[{Liu et~al.(2022{\natexlab{d}})Liu, Hu, Kang, Luo, and Fan}]{liu2022interactformer}
Liu, Y.; Hu, J.; Kang, X.; Luo, J.; and Fan, S. 2022{\natexlab{d}}.
\newblock Interactformer: Interactive transformer and CNN for hyperspectral image super-resolution.
\newblock \emph{IEEE Transactions on Geoscience and Remote Sensing}, 60: 1--15.

\bibitem[{Mao et~al.(2023)Mao, Jiang, Chen, and Li}]{mao2023disc}
Mao, Y.; Jiang, L.; Chen, X.; and Li, C. 2023.
\newblock DisC-Diff: Disentangled Conditional Diffusion Model for Multi-Contrast MRI Super-Resolution.
\newblock \emph{arXiv preprint arXiv:2303.13933}.

\bibitem[{Nichol and Dhariwal(2021)}]{pmlr-v139-nichol21a}
Nichol, A.~Q.; and Dhariwal, P. 2021.
\newblock Improved Denoising Diffusion Probabilistic Models.
\newblock In Meila, M.; and Zhang, T., eds., \emph{Proceedings of the 38th International Conference on Machine Learning}, volume 139 of \emph{Proceedings of Machine Learning Research}, 8162--8171. PMLR.

\bibitem[{Rombach et~al.(2022)Rombach, Blattmann, Lorenz, Esser, and Ommer}]{rombach2022high}
Rombach, R.; Blattmann, A.; Lorenz, D.; Esser, P.; and Ommer, B. 2022.
\newblock High-resolution image synthesis with latent diffusion models.
\newblock In \emph{Proceedings of the IEEE/CVF conference on computer vision and pattern recognition}, 10684--10695.

\bibitem[{Saharia et~al.(2022)Saharia, Ho, Chan, Salimans, Fleet, and Norouzi}]{saharia2022image}
Saharia, C.; Ho, J.; Chan, W.; Salimans, T.; Fleet, D.~J.; and Norouzi, M. 2022.
\newblock Image super-resolution via iterative refinement.
\newblock \emph{IEEE Transactions on Pattern Analysis and Machine Intelligence}, 45(4): 4713--4726.

\bibitem[{Saharia et~al.(2023)Saharia, Ho, Chan, Salimans, Fleet, and Norouzi}]{9887996}
Saharia, C.; Ho, J.; Chan, W.; Salimans, T.; Fleet, D.~J.; and Norouzi, M. 2023.
\newblock Image Super-Resolution via Iterative Refinement.
\newblock \emph{IEEE Transactions on Pattern Analysis and Machine Intelligence}, 45(4): 4713--4726.

\bibitem[{Simonyan and Zisserman(2014)}]{brusilovsky:simonyan2014very}
Simonyan, K.; and Zisserman, A. 2014.
\newblock {Very deep convolutional networks for large-scale image recognition}.
\newblock \emph{arXiv preprint arXiv:1409.1556}.

\bibitem[{Thai and Healey(2002)}]{2002Invariant}
Thai, B.; and Healey, G. 2002.
\newblock Invariant subpixel material detection in hyperspectral imagery.
\newblock \emph{IEEE Transactions on Geoscience and Remote Sensing}, 40(3): 599--608.

\bibitem[{Wang et~al.(2022)Wang, Hu, Jiang, and Ma}]{9930808}
Wang, X.; Hu, Q.; Jiang, J.; and Ma, J. 2022.
\newblock A Group-Based Embedding Learning and Integration Network for Hyperspectral Image Super-Resolution.
\newblock \emph{IEEE Transactions on Geoscience and Remote Sensing}, 60: 1--16.

\bibitem[{Wang, Ma, and Jiang(2022)}]{9380508}
Wang, X.; Ma, J.; and Jiang, J. 2022.
\newblock Hyperspectral Image Super-Resolution via Recurrent Feedback Embedding and Spatial–Spectral Consistency Regularization.
\newblock \emph{IEEE Transactions on Geoscience and Remote Sensing}, 60: 1--13.

\bibitem[{Yokoya and Iwasaki(2016)}]{unknown}
Yokoya, N.; and Iwasaki, A. 2016.
\newblock Airborne hyperspectral data over Chikusei.

\bibitem[{Zhou et~al.(2023)Zhou, Sheng, Fan, Ye, He, Wang, and Chen}]{zhou2023hyperspectral}
Zhou, J.; Sheng, J.; Fan, J.; Ye, P.; He, T.; Wang, B.; and Chen, T. 2023.
\newblock When Hyperspectral Image Classification Meets Diffusion Models: An Unsupervised Feature Learning Framework.
\newblock \emph{arXiv preprint arXiv:2306.08964}.

\end{thebibliography}

\end{document}